\documentclass[lettersize,journal]{IEEEtran}
\usepackage{amsmath,amsfonts}
\usepackage{algorithmic}
\usepackage{algorithm}
\usepackage[caption=false,font=normalsize,labelfont=sf,textfont=sf]{subfig}
\usepackage{textcomp}
\usepackage{stfloats}
\usepackage{url}
\usepackage{verbatim}
\usepackage{graphicx}
\usepackage{cite}
\usepackage{arydshln}
\usepackage{booktabs}
\usepackage{multirow}
\usepackage{float}
\usepackage{hyperref} 
\usepackage[table,xcdraw]{xcolor} % Enable color in tables
\begin{document}

\title{FGAseg: Fine-Grained Pixel-Text Alignment for Open-Vocabulary Semantic Segmentation}

\author{
Bingyu Li,
Da Zhang,~\IEEEmembership{Student Member,~IEEE,} 
Zhiyuan Zhao,
Junyu Gao,~\IEEEmembership{Member,~IEEE,} \\
and
Xuelong Li\(^*\),~\IEEEmembership{Fellow,~IEEE}
  \thanks{*Corresponding author: Xuelong Li.}
  \thanks{Bingyu Li is with the Department of Electronic Engineering and Information Science, University of Science and Technology of China, He Fei 230026, P. R. China and the Institute of Artificial Intelligence (TeleAI), China Telecom, P. R. China.(E-mail: libingyu0205@mail.ustc.edu.cn).}
  \thanks{Zhiyuan Zhao and Xuelong Li are with  the Institute of Artificial Intelligence (TeleAI), China Telecom, P. R. China. (E-mail: tuzixini@163.com; xuelong\_li@ieee.org).}
  \thanks{Da Zhang and Junyu Gao are with the School of Artificial Intelligence, OPtics and ElectroNics (iOPEN), Northwestern Polytechnical University, Xi'an 710072, China and with the Institute of Artificial Intelligence (TeleAI), China Telecom, P. R. China. (E-mail: zhangda1018@126.com; gjy3035@gmail.com).}
  }



\maketitle

\begin{abstract}
Open-vocabulary segmentation aims to identify and segment specific regions and objects based on text-based descriptions. A common solution is to leverage powerful vision-language models (VLMs), such as CLIP, to bridge the gap between vision and text information. However, VLMs are typically pretrained for image-level vision-text alignment, focusing on global semantic features. In contrast, segmentation tasks require fine-grained pixel-level alignment and detailed category boundary information, which VLMs alone cannot provide. As a result, information extracted directly from VLMs can't meet the requirements of segmentation tasks.
To address this limitation, we propose FGAseg, a model designed for fine-grained pixel-text alignment and category boundary supplementation. 
The core of FGAseg is a Pixel-Level Alignment module that employs a cross-modal attention mechanism and a text-pixel alignment loss to refine the coarse-grained alignment from CLIP, achieving finer-grained pixel-text semantic alignment.
Additionally, to enrich category boundary information, we introduce the alignment matrices as optimizable pseudo-masks during forward propagation and propose Category Information Supplementation module. These pseudo-masks, derived from cosine and convolutional similarity, provide essential global and local boundary information between different categories.
By combining these two strategies, FGAseg effectively enhances pixel-level alignment and category boundary information, addressing key challenges in open-vocabulary segmentation. 
Extensive experiments demonstrate that FGAseg outperforms existing methods on open-vocabulary semantic segmentation benchmarks. The code is \href{https://github.com/LiBingyu01/FGA-seg}{here}.
\end{abstract}
\section{Introduction}
\label{sec:intro}
Semantic segmentation is a fundamental task in computer vision, aiming to achieve pixel-level category prediction \cite{chen2017deeplab}. Its precise predictive capabilities play a crucial role in fields such as autonomous driving \cite{zhang2023cmx} and remote sensing \cite{zhang2024integrating}. Traditional semantic segmentation models rely on manually annotated datasets, which are typically limited in scope to a few dozen categories. This reliance not only restricts their ability to recognize unseen categories during training but also hinders their broader applicability due to constrained generalization capabilities. Moreover, the high resource costs associated with manual annotation datasets further limit the diversity of categories included in these datasets, exacerbating the issue \cite{xie2024sed}.

The recent rise of vision-language models (VLMs) has prompted a re-evaluation of scalability in semantic segmentation \cite{bucher2019zero}. Specifically, models like CLIP \cite{radford2021learning} and ALIGN \cite{jia2021scaling}, trained through cross-modal pretraining, exhibit robust vision-text alignment capabilities. This cross-modal alignment is crucial for aligning vision information with text-based descriptions in segmentation models, effectively expanding the number and range of categories that can be segmented \cite{li2022language}. 

\begin{figure}
\centering
\includegraphics[width=\linewidth]{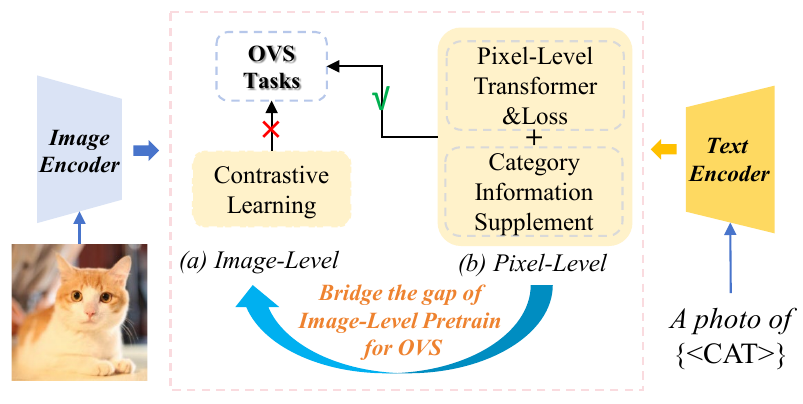}
\caption{\textbf{Comparison of Image-Level Pretraining and Pixel-Level Alignment.} (a) Image-Level Pretraining aligns image and text embeddings via contrastive learning, while (b) Pixel-Level Alignment incorporates a pixel-level transformer, alignment loss and category information supplement to achieve finer-grained alignment, bridging the gap for open-vocabulary segmentation (OVS).}
\label{fig:fig_1}
\end{figure}

By leveraging VLMs in semantic segmentation tasks, researchers have introduced open-vocabulary segmentation (OVS), enabling models to overcome the limitations of specific training data and achieve segmentation of arbitrary categories \cite{xu2023side}. However, directly applying VLMs to semantic segmentation presents several challenges, particularly due to the gap between the pretrained vision-text alignment at the image level in VLMs and the finer-grained region-text and pixel-text alignment required for semantic segmentation.
To bridge this gap, various solutions have been proposed, generally categorized into two-stage and single-stage models. Two-stage approaches, such as \cite{liang2023open, xu2022simple}, use mask generators to produce segmentation masks that serve as intermediate information for downstream models. These methods often rely on pretrained backbones to classify pixels based on the generated masks \cite{zhou2022extract}. However, the accuracy of these masks is frequently suboptimal, which limits their applicability. In contrast, single-stage models like FC-CLIP \cite{yu2023convolutions} and its variants \cite{jiao2025collaborative} integrate a trainable mask generator directly within the VLM framework, producing segmentation prompts that enhance efficiency and effectiveness. Other methods, such as CAT-SEG \cite{cho2024cat} and SED \cite{xie2024sed}, leverage similarity matrices as implicit pseudo-masks, using high-similarity prompts during forward propagation to improve segmentation performance without explicit mask generation. However, although this method provides potential pseudo-masks as boundary cues, it still lacks fine-grained boundary masks for capturing localized details.
Despite these advancements, achieving fine-grained, pixel-level alignment between vision and text remains a challenge. Current models rarely address this need while simultaneously preserving rich category boundary information, which is critical for precise semantic segmentation. We propose that effective open-vocabulary semantic segmentation (OVS) requires both fine-grained pixel-level alignment and the retention of detailed category boundary information.

To achieve this goal, we propose the FGASeg framework, which enables pixel-text alignment and provides the necessary boundary information. As shown in Fig. \textcolor{red}{\ref{fig:fig_1}}, this framework primarily consists of two key components: the Pixel-Level Alignment module and Category Supplementation Propagation module. The Pixel-Level Alignment module is a bidirectional fine-grained alignment module. For pixel-text alignment, we design the Pixel-Text Alignment Transformer (P2Tformer) to convert vision-text alignment into fine-grained pixel-text alignment. For text-pixel alignment, we introduce the Text-Pixel Alignment Loss (T2Ploss), which guides the vision encoder to gain text-pixel alignment while avoiding deviation from the pretrained image-text alignment.
Further, to provide essential boundary information for precise mask prediction, we propose the Category Supplementation Propagation module, which uses alignment matrices as pseudo-masks for features propagation. Specifically, we calculate both global and local feature similarities by employing cosine and convolution-based similarity matrices. During forward propagation, features with high similarity are treated as potential masks and participate in information propagation as category information providers. These two similarity calculation methods balance local and global alignment, thereby enhancing segmentation performance.

Overall, our contributions are as follows:
\begin{itemize}
    \item We introduce FGASeg, a framework tailored for open-vocabulary segmentation that enables pixel-level alignment between vision and text, addressing the gap in VLMs' image-level alignment.
    \item Our Pixel-Text Alignment Transformer (P2Tformer) and Text-Pixel Alignment Loss (T2Ploss) work together to transform VLMs' coarse image-text alignment into precise pixel-text alignment for improved segmentation.
    \item We propose a Category Supplementation Propagation module, utilizing cosine and convolution-based similarity matrices as pseudo-masks to enrich category boundary cues and enhance segmentation accuracy.
    \item Our model, which incorporates pixel-level alignment and essential category boundary information, demonstrates satisfactory performance across multiple commonly used datasets.
\end{itemize}

The overall structure of the paper is as follows: section \ref{related_work} introduces related work, section \ref{method} presents the details of the model framework, section \ref{experiments_results} provides comprehensive experiments and visualizations, and section \ref{sec:conclusion} concludes with a summary and future outlook.

\begin{figure*}
\centering
\includegraphics[width=\linewidth]{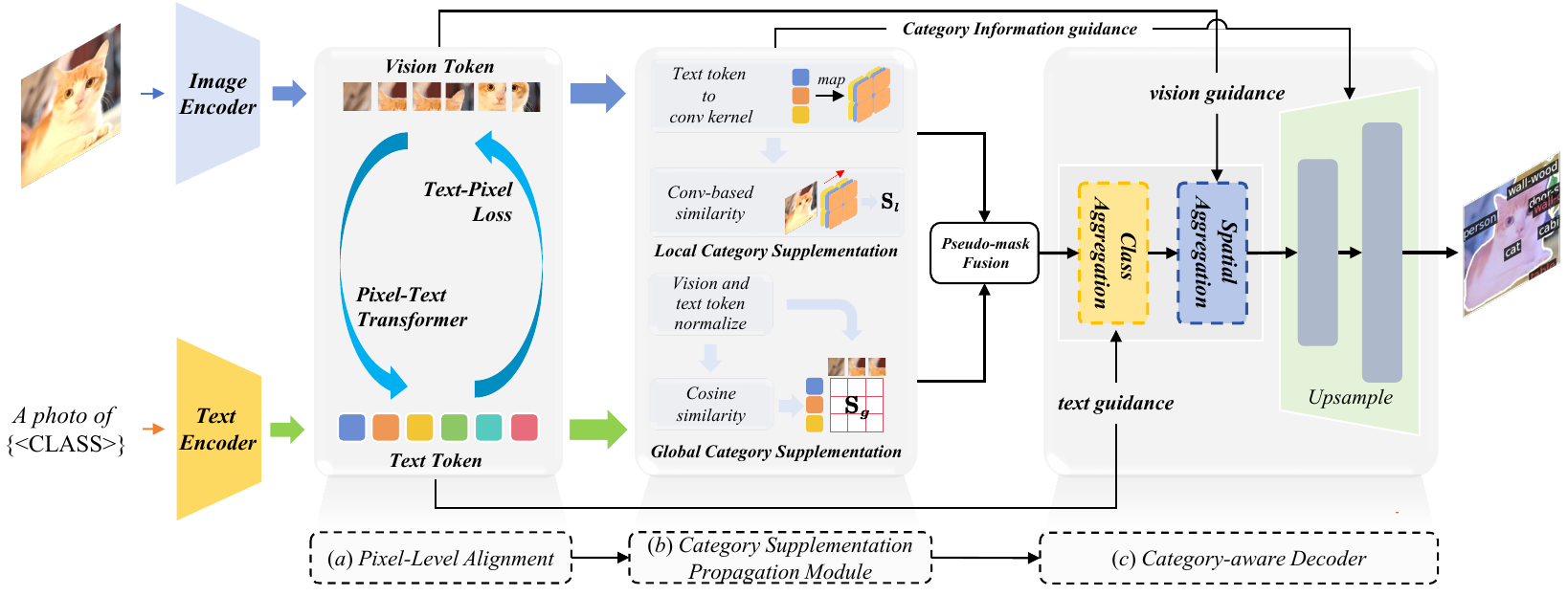}
\caption{\textbf{Overall architecture of FGAseg.} (a) Pixel-Level Alignment Module: The P2Tformer aligns tokens in a pixel-text manner, while the T2Ploss enforces precise text-pixel alignment through local alignment. (b) Global Category Supplementation (GCS) and Local Category Supplementation (LCS) provide category boundary information as pseudo-masks for guidance. (c) Global and Local Category Supplementation Propagation incorporates GCS and LCS as pseudo-masks into pixel-level classification.}
\label{fig:fig_2}
\end{figure*}

\section{Related Work}
\label{related_work}
\subsection{Semantic Segmentation}
Semantic segmentation, which aims to achieve pixel-level classification, has made significant progress \cite{li2024stitchfusion}. 
In the past, many classical methods made significant progress in the field of semantic segmentation. However, it is the application of convolutional structures that has propelled semantic segmentation into a new era.
Early breakthroughs were mainly driven by fully convolutional networks (FCNs), which served as end-to-end models for semantic segmentation \cite{long2015fully}. Subsequently, models like SegNet \cite{badrinarayanan2017segnet} and U-net \cite{ronneberger2015u}, evolved from FCNs, gained widespread attention.
Another significant advancement is the introduction of ResNet\cite{he2016deep}, which, pre-trained on ImageNet, has demonstrated strong performance in transferring to downstream tasks. In the field of semantic segmentation, ResNet has been widely adopted as a feature encoder, serving as the foundation for many groundbreaking segmentation models\cite{lin2018deeptongue}. With the use of ResNet, researchers have been able to focus more on the intrinsic characteristics of the segmentation task itself. For instance, the Deeplab series \cite{chen2017deeplab}, known for its pyramid structure, has gained significant attention for its ability to facilitate multi-scale information extraction and fusion.

In recent years, Vision Transformers (ViT) \cite{dosovitskiy2020image} have enhanced the capability of semantic segmentation models by leveraging attention mechanisms to establish global dependencies in images \cite{na2024switching, zhao2024improving}. Based on ViT, numerous semantic segmentation paradigms have been proposed, expanding their capabilities \cite{li2024u3m, hu2024contrastive}. The development of these models is closely related to the detailed and dense nature of semantic segmentation tasks\cite{xie2024multi, zhu2024saswot}. Unlike classification tasks, which only require the extraction and estimation of overall image features, semantic segmentation demands pixel-level feature localization, shaping, and classification\cite{li2024stitchfusion}. Therefore, more meticulous feature extraction and discrimination are necessary. The academic community has made significant contributions in this regard.
Notable works, such as MaskFormer \cite{cheng2021per} and Mask2Former \cite{cheng2022masked}, have unified pixel-level classification and mask classification to achieve superior performance. SegFormer, on the other hand, addresses the resolution issue by eliminating positional encodings and providing multi-resolution features \cite{xie2021segformer}. 

\subsection{Open Vocabulary Segmentation}  
Although semantic segmentation models have developed extensively, they rely on specific datasets for training, which are often limited to particular scenes and categories \cite{yu2023convolutions}. This limitation affects the scalability and generative capabilities of traditional semantic segmentation models .  
Fortunately, with the rise of pretrained Vision-Language Models (VLMs) \cite{radford2021learning}, semantic segmentation has embraced a new paradigm, where aligning vision information with text-based descriptions allows segmentation models to handle a broader range of categories \cite{xu2023side,cho2024cat}. However, VLMs such as CLIP and ALIGN are pretrained in a vision-text manner at image-level \cite{liang2023open, han2023open}, while semantic segmentation requires fine-grained pixel-text alignment, therefore, directly applying VLMs to downstream segmentation tasks often results in a significant gap.  

To bridge the gap between vision-language alignment and the demands of semantic segmentation, various methodologies have emerged, broadly categorized into two-stage and single-stage approaches. Two-stage frameworks, such as those proposed in \cite{liang2023open, xu2022simple}, employ mask generators to produce segmentation masks that supplement downstream segmentation models. Similarly, approaches like \cite{zhou2022extract} utilize masks extracted from pretrained models like CLIP, incorporating additional backbones to classify pixels based on the generated masks. Despite their structured pipeline, the limited accuracy of mask generation in these methods often undermines their general applicability, prompting an increased focus on single-stage models that are both simpler and more efficient. 
The inaccuracies in mask generation within two-stage frameworks have motivated the development of integrated solutions. Single-stage models, such as FC-CLIP \cite{yu2023convolutions} and its collaborative extension \cite{jiao2025collaborative}, embed trainable mask generators directly into the CLIP architecture, allowing the generated masks to act as segmentation prompts. Moving beyond explicit mask generation, advanced techniques like CAT-Seg \cite{cho2024cat} and SED \cite{xie2024sed} exploit similarity matrices as pseudo-masks during forward propagation, offering high-similarity cues that enhance segmentation quality while avoiding the challenges of mask generation.
Recent studies have further expanded the landscape of open-vocabulary segmentation through innovative mechanisms. For instance, side adapters for efficient feature fusion have been introduced in \cite{xu2023side}, while frequency-domain modules for robust generalization have been proposed in \cite{xu2024generalization}. Moreover, \cite{shan2024open} demonstrated adaptive integration of outputs from SAM and CLIP, achieving significant gains in segmentation performance. These advancements collectively illustrate the evolving strategies in integrating vision-language alignment with segmentation tasks, emphasizing efficiency and adaptability.
\section{Method}
\label{method}
\subsection{Preliminary}
\subsubsection{Problem Definition}
Open Vocabulary Segmentation (OVS) aims to segment an image into semantically meaningful regions without being constrained by a fixed set of predefined categories. Unlike traditional segmentation tasks, which rely on a closed set of known classes, OVS seeks to generalize to unseen categories by leveraging information such as text-based descriptions. The problem can be formally defined as follows:

Given a batch of images \( \mathbf{I} = \{I_1, I_2, \ldots, I_B\} \) and a set of textual descriptions \( \mathbf{T} = \{t_1, t_2, \ldots, t_T\} \), where \(B\) is the number of images of a batch and \(T\) is the number of defined categories. The goal is to generate a segmentation map \( \mathbf{Y} \) where each pixel \( p \in \mathbf{I} \) is assigned a label from an open vocabulary derived from \( \mathbf{T} \). The segmentation map \( \mathbf{Y} \) should accurately reflect the semantic regions corresponding to the descriptions in \( \mathbf{T} \), even if the categories are absent in the training data.
Formally, the task is to learn a mapping:\(f: \mathbf{I} \times \mathbf{T} \rightarrow \mathbf{Y}\) such that \( f(\mathbf{I}, \mathbf{T}) = \mathbf{Y} \), where \( \mathbf{Y} \) is a pixel-wise labeling of \( \mathbf{I} \) with labels from an open vocabulary.

\subsubsection{Review of Vision-Language Models}

Vision-Language Models (VLMs) aim to align vision and text representations in image-level.
The typical architecture consists of two main components: an image encoder and a text encoder. The image encoder processes the image \( \mathbf{I} = \{I_1, I_2, \ldots, I_B\} \) to extract a vision feature vector \( \mathbf{V}_\mathbf{I} \in \mathbb{R}^{B \times C \times H \times W} \). The text encoder processes a sequence of words \( \mathbf{T} = \{t_1, t_2, \ldots, t_T\} \) to produce a corresponding feature vector \( \mathbf{V}_\mathbf{T} \in \mathbb{R}^{B \times T \times d}\) (repeated \(B\) times), \(d=C\), where \( B \) represents the batch size, \( C \) denotes the number of feature channels, \( H \) and \( W \) are the spatial dimensions of the vision feature, \( T \) refers to the number of defined categories, and \( d \) is the text feature dimension, which equals \( C \) in this paper..
\begin{equation}
\mathbf{V}_\mathbf{I} = \text{ImageEncoder}(I_1, I_2, \ldots, I_B)
\end{equation}
\begin{equation}
\mathbf{V}_\mathbf{T} = \text{TextEncoder}(t_1, t_2, \ldots, t_T)
\end{equation}

The goal is to maximize the cosine similarity between the vision and text embeddings for matching pairs and minimize it for non-matching pairs. The cosine similarity is given by:
\begin{equation}
\mathcal{S}(\mathbf{V}_\mathbf{I}, \mathbf{V}_\mathbf{T}) = \frac{\mathbf{V}_\mathbf{I} \cdot \mathbf{V}_\mathbf{T}}{\|\mathbf{V}_\mathbf{I}\| \|\mathbf{V}_\mathbf{T}\|}
\end{equation}

The loss function, often a contrastive loss, can be expressed as:
\begin{equation}
\mathcal{L} = - \frac{1}{N} \sum_{i=1}^{N}\log \frac{\exp(\mathcal{S}_{i,i}/\tau)}{\sum_{j=1}^{N} \exp(\mathcal{S}_{i,j}/\tau) + \sum_{j=1}^{N} \exp(\mathcal{S}_{j,i}/\tau)},
\end{equation}
where \(\tau\) is a temperature parameter that controls the concentration level of the distribution. $\mathcal{S}_{i,j}$ is the similarity matrix between the $i$-th sample and the $j$-th text sample. $N$ is the number of samples.

\begin{figure}[ht]
\centering
\includegraphics[width=0.9\linewidth]{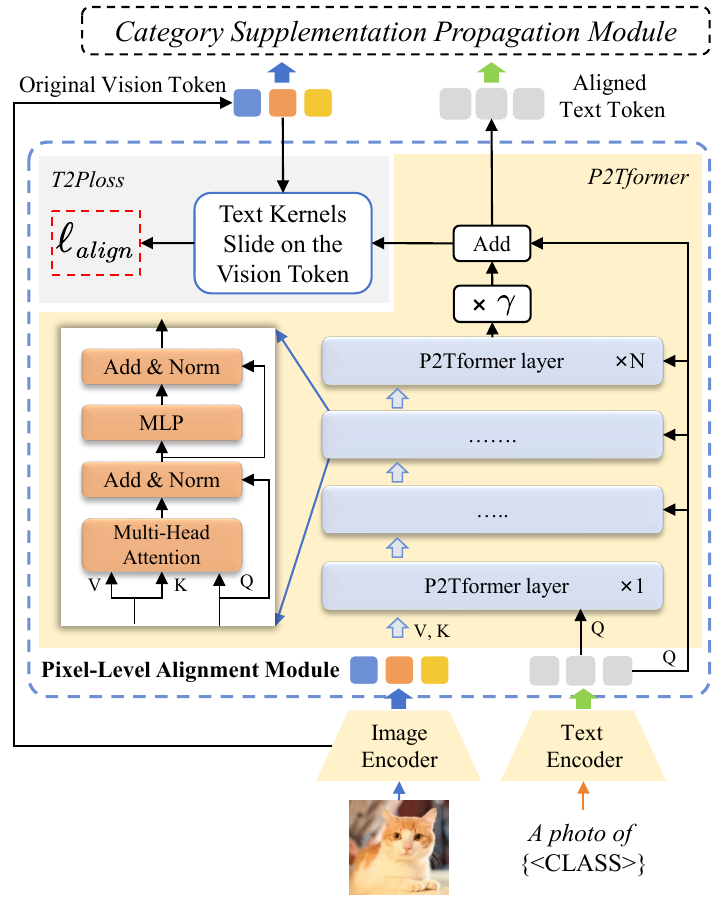}
\caption{\textbf{Pixel-Level Alignment Module.} This module refines pixel-text alignment by using multi-head attention and MLP layers across multiple P2Tformer layers. Vision and text tokens are processed to capture cross-modal correlations, and the result is scaled by a learnable parameter $\gamma$, to balance the modalities. Aligned text tokens slide over vision tokens, creating alignment matrices for computing alignment loss ($\mathcal{L}_{\text{align}}$) via T2Ploss, enhancing segmentation precision by preserving category boundaries. This fine-grained pixel-text alignment enhances the model’s ability to capture category boundaries.}
\label{fig:fig_3}
\end{figure}

Although Vision-Language Models (VLMs) have been trained on large-scale datasets and successfully aligned vision and text features, their training paradigms are predominantly image-level focused.

\subsection{Pixel-Level Alignment Module}
To achieve pixel level alignment for segmentation task, we propose Pixel-Level Aligment Module, Fig. \ref{fig:fig_3} illustrates the process of the Pixel-Level Alignment Module.
\subsubsection{Pixel-Text Alignment Transformer} 
The Pixel-Text Alignment Transformer is designed to achieve pixel-level cross-modal alignment between image features and semantic textual descriptions. This module, referred to as P2Tformer, aims to transfer and align vision information from the vision domain to the text-based space in pixel-text manner.

The P2Tformer module takes the vision features $\mathbf{V}_\mathbf{I} \in \mathbb{R}^{B \times C \times H \times W}$ and the text features $\mathbf{V}_\mathbf{T} \in \mathbb{R}^{B \times T \times d}$ as inputs.

Firstly, a sinusoidal positional encoding $\mathbf{P}$ is added to the flattened vision features. The encoded image features are denoted by:
\(\mathbf{P}_{\text{flat}} = \text{Flatten}(\text{PE}(\mathbf{V}_\mathbf{I})), \)where $\mathbf{P}_{\text{flat}} \in \mathbb{R}^{B \times (H \cdot W) \times d}$, $\text{Flatten}$ denotes flattening the height and width dimensions, and $\text{PE}(\cdot)$ represents the sinusoidal positional embedding function.
Then, Cross-attention layers are used to align the vision features $\mathbf{V}_\mathbf{I}$ with the repeated text embeddings $\mathbf{V}_\mathbf{T}$ through multiple layers (denoted by $N$ in Fig. \textcolor{red}{\ref{fig:fig_3}}). For $i$-th P2Tformer layer, the output text features $\mathbf{V}_{\mathbf{T},\text{out}}^{i}$ are computed via cross-attention, which is represented by:
\begin{equation}
\mathbf{V}_{\mathbf{T},\text{out}}^{i} = \text{CrossAttention}(\mathbf{V}_{\mathbf{T},\text{out}}^{i-1}, \mathbf{V}_{\mathbf{I}}^\text{flat} + \mathbf{P}_{\text{flat}})
\end{equation}
where, $\mathbf{V}_{\mathbf{T},\text{out}}^{1}$ is $\mathbf{V}_{\mathbf{T}}$, $\mathbf{V}_{\mathbf{I}}^\text{flat}$ represents the flattened image features and $\mathbf{P}_{\text{flat}}$ represents the positional encoding added to them. The cross-attention helps the text features attend to relevant vision elements in the Pixel2Text manner.

The final text embeddings $\mathbf{V}_\mathbf{T}^{\text{emb}}$ are obtained by combining the original text embeddings $\mathbf{V}_\mathbf{T}$ with the output of the cross-attention mechanism scaled by a learnable parameter $\gamma$:
\begin{equation}
\mathbf{V}_\mathbf{T}^{\text{emb}} = \mathbf{V}_\mathbf{T} + \gamma \cdot \mathbf{V}_{\mathbf{T},\text{out}}^{N}
\end{equation}
where $\gamma$ is a trainable parameter that controls the contribution of the cross-attended vision information.

The P2Tformer achieves pixel-level alignment between vision and text using cross-attention mechanisms, capturing fine-grained vision-semantic alignment.
\subsubsection{Text-Pixel Alignment Loss}
To align vision and text-based information bidirectionally, we additionally introduce the Text-Pixel Alignment Loss (T2Ploss) in this subsection.

The vision input, denoted as \(\mathbf{V}_\mathbf{I}\), is passed through a projection network to gain a feature map with more detailed category information. The resulting feature map \(\mathbf{V}_\mathbf{I}' \in \mathbb{R}^{B \times \frac{C}{2} \times 4H \times W}\).
The text input represented as \(\mathbf{V}_\mathbf{T}\) is projected using a linear transformation. This produces convolutional kernels \(\mathbf{W} \in \mathbb{R}^{B \times T \times \frac{d}{2} \times K \times K}\) and biases \(\mathbf{b} \in \mathbb{R}^{B \times T} \). To calculate the local alignment between the vision and text features, the projected vision features \(\mathbf{V}_\mathbf{I}'\) is convolved with the generated kernels \(\mathbf{W}\), along with adding biases \(\mathbf{b}\):
\begin{equation}
\mathbf{O}_i = \text{Conv2D}(\mathbf{V}_{\mathbf{I},i}', \mathbf{W}_i) + \mathbf{b}_i
\end{equation}
where \(\mathbf{O}_i\) represents the similarity map for the \(i\)-th batch, resulting in an output tensor \(\mathbf{O} \in \mathbb{R}^{B \times T \times 4H \times 4W}\).

The \(\mathbf{O}\) is transposed and upsampled to align its dimensions with the mask \( \mathbf{M}\). The resulting local alignment feature is represented as \(\mathbf{O}_{\text{align}} \in \mathbb{R}^{B \times T \times H \times W}\).

The local alignment features and the ground truth are used to compute the alignment loss, ensuring the model aligns the local text-pixel representations. To compute the loss, we use a binary mask \(\mathbf{M} \in \mathbb{R}^{B \times H \times W}\). The alignment loss is calculated using a mean squared error (MSE) loss between the downsampled similarity output \(\mathbf{O}_{\text{align}}\) and the ground-truth mask \(\mathbf{M}\):
\begin{equation}
\mathcal{L}_{\text{align}} = \frac{1}{B} \sum_{i=1}^{B} \sum_{j=1}^{H} \sum_{k=1}^{W} \left( \mathbf{O}_{\text{align}, i, j, k} - \mathbf{M}_{i, j, k} \right)^2
\end{equation}

The T2Ploss \(\mathcal{L}_{\text{align}}\) is crucial in guiding the model to align the text-based descriptions with the local vision features highlighted by the mask, thereby enhancing the alignment in pixel-level manner.

\subsection{Category Supplementation Propagation Module}
\begin{figure}[ht]
\centering
\includegraphics[width=0.98\linewidth]{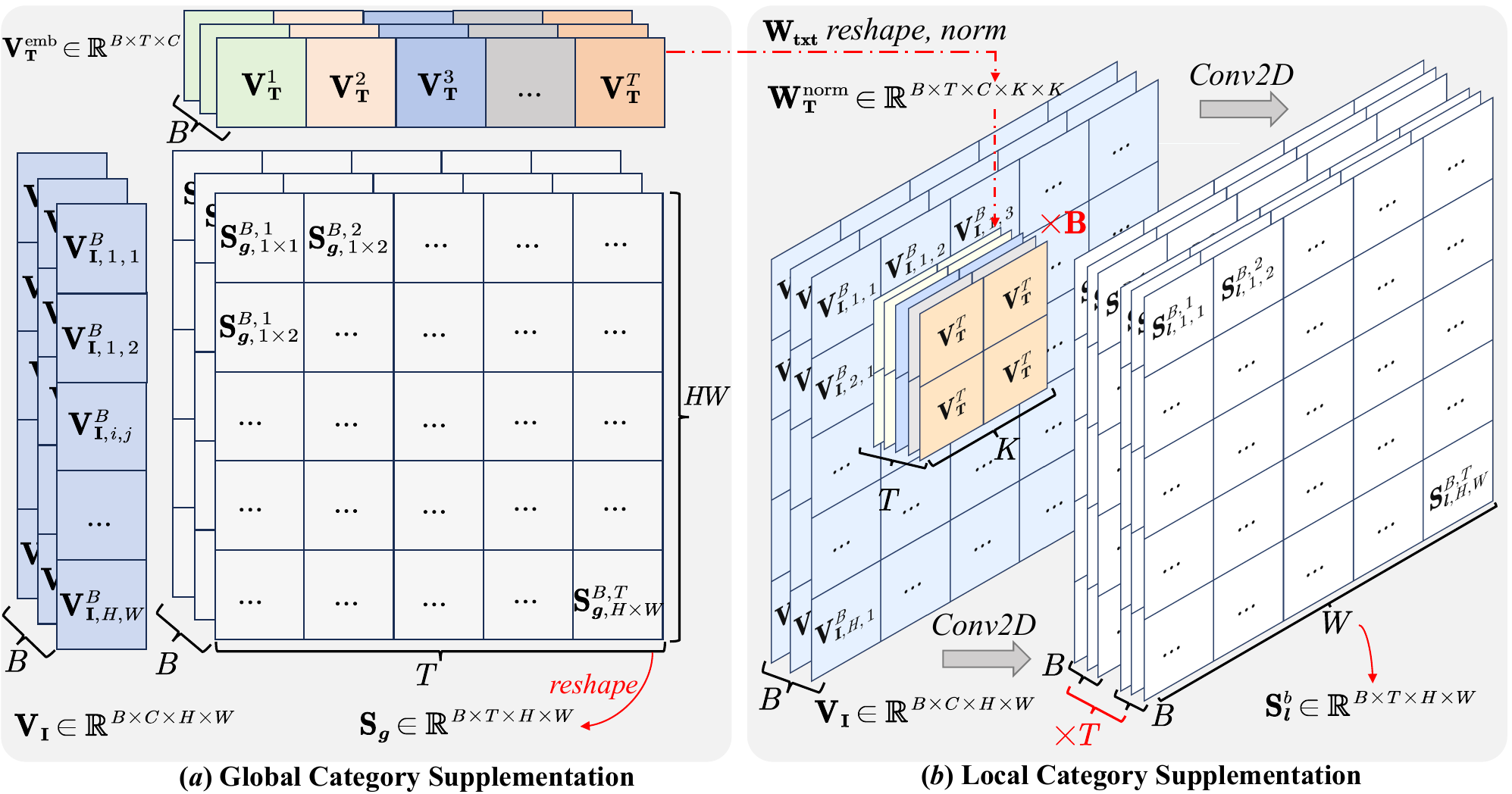}
\caption{\textbf{Illustration of (a) Global Category Supplementation(GCS) and (b) Local Category Supplementation (LCS).} (a) Global category information is obtained by computing the cosine similarity,(b) Local category information is obtained by applying a text convolutional kernel to slide over the image features}
\label{fig:fig_4}
\end{figure}

In this section, we explore methods for Category Supplementation  Propagation, which primarily include global and local category information supplementation modules, demonstrated in Fig. \textcolor{red}{\ref{fig:fig_4}} (a) and (b), respectively.

\subsubsection{Global Category Supplementation(GCS)}
We employ cosine similarity to capture global category boundary information. Cosine similarity between features \( \mathbf{V}_\mathbf{T}^{\text{emb}} \) and \( \mathbf{V}_\mathbf{I} \) is defined as:
\begin{equation}
\mathcal{S}_{g} = \frac{\mathbf{V}_\mathbf{T}^{\text{emb}} \cdot \mathbf{V}_\mathbf{I}}{\|\mathbf{V}_\mathbf{T}^{\text{emb}}\| \|\mathbf{V}_\mathbf{I}\|},
\end{equation}
where, \( \mathbf{V}_\mathbf{T}^{\text{emb}} \cdot \mathbf{V}_\mathbf{I} \) is the dot product of vectors \( \mathbf{V}_\mathbf{T}^{\text{emb}} \) and \( \mathbf{V}_\mathbf{I} \). \( \|\mathbf{V}_\mathbf{T}^{\text{emb}}\| \) and \( \|\mathbf{V}_\mathbf{I}\| \) are the magnitudes of vectors \( \mathbf{V}_\mathbf{T}^{\text{emb}} \) and \( \mathbf{V}_\mathbf{I} \).
Cosine similarity reflects the similarity between each pixel block and text. During subsequent forward propagation, the cosine similarity with high similarity are emphasized, while areas with low similarity are suppressed.

\subsubsection{Local Category Supplementation (LCS)}
Like GCS, the core idea of LCS is to highlight areas with high similarity for subsequent propagation while suppressing those with low similarity.
However, LCS employs a convolution-based similarity to capture more local detailed information.
The forward propagation of the network takes vision features $\mathbf{V}_\mathbf{I}$ and text features $\mathbf{V}_\mathbf{T}^{\text{emb}}$ as inputs. The text features $\mathbf{V}_\mathbf{T}^{\text{emb}}$ are projected to match the dimensions of the vision features using a linear transformation, such that:
\(\mathbf{V}_\mathbf{T}^{\text{proj}} = \mathbf{W}_{\text{txt}} \mathbf{V}_\mathbf{T}^{\text{emb}},\)
where $\mathbf{W}_{\text{txt}}\in \mathbb{R}^{d \times C\cdot K \cdot K}$ is the linear layer parameter matrix. The output text features $\mathbf{V}_\mathbf{T}^{\text{proj}}$ are then used to generate the weight $\mathbf{W}_\mathbf{T}^{\text{proj}} \in \mathbb{R}^{B \times T \times C \times K \times K}$ and bias $\mathbf{b}_\mathbf{T}^{\text{proj}} \in \mathbb{R}^{B \times T}$ for the convolutional operations.
To normalize each kernel, the projected kernel \(\mathbf{W}_\mathbf{T}^{\text{proj}}\) is reshaped using pixel shuffling, then softmax is applied along the channel dimension to highlight areas with higher similarity:
$\mathbf{W}_\mathbf{T}^{\text{norm}} = \text{Softmax}(\mathbf{W}_\mathbf{T}^{\text{proj}}).$

The reshaped mask $\mathbf{W}_\mathbf{T}^{\text{norm}}$ is applied to calculate local similarities between the vision and text features. Next, the local correlation similarities are computed through depthwise convolution between vision features $\mathbf{V}_\mathbf{I}$ and the normalized kernel weights. The convolution is performed using the following formulation:
\begin{equation}
\mathcal{S}_{l}= \text{Conv2D}(\mathbf{V}_\mathbf{I}, \mathbf{W}_\mathbf{T}^{\text{norm}}) + \mathbf{b}_{\text{T}}^{\text{proj}},
\end{equation}
where, $K$ represents the kernel size, $\mathbf{W}_\mathbf{T}^{\text{norm}}$ is the weight tensor after projection, and $\mathbf{b}_{\text{T}}^{\text{proj}}$ is the bias term. The aligned output $\mathcal{S}_{l}(\mathbf{V}_\mathbf{T}^{\text{emb}}, \mathbf{V}_\mathbf{I})$ represents the local pixel-wise correlation between the image and text, highlighting regions that are more semantically similar with the category boundary feature.

\subsubsection{Global and Local Category Supplementation Propagation}
As shown in Fig. \textcolor{red}{\ref{fig:fig_2}}(c), in the forward propagation process, GCS $\mathcal{S}_{g}$ and LCS $\mathcal{S}_{l}$ are first fused. The fused feature represents both global features and fine-grained local boundary features. This fused map acts as a pseudo-mask for forward propagation. During propagation, the pseudo-mask is integrated with the hierarchical features of the previous CLIP model (vision guidance and text guidance in Fig. \textcolor{red}{\ref{fig:fig_2}}) in the form of a pseudo-mask. It undergoes class aggregation and spatial aggregation to model the features' categorical and spatial characteristics \cite{cho2024cat}. Class aggregation consists of a two-layer cross-attention mechanism that uses text guidance as the query, while spatial aggregation is a Swin Transformer module that fuses pseudo-masks with vision guidance.

\begin{figure}[ht]
\centering
\includegraphics[width=\linewidth]{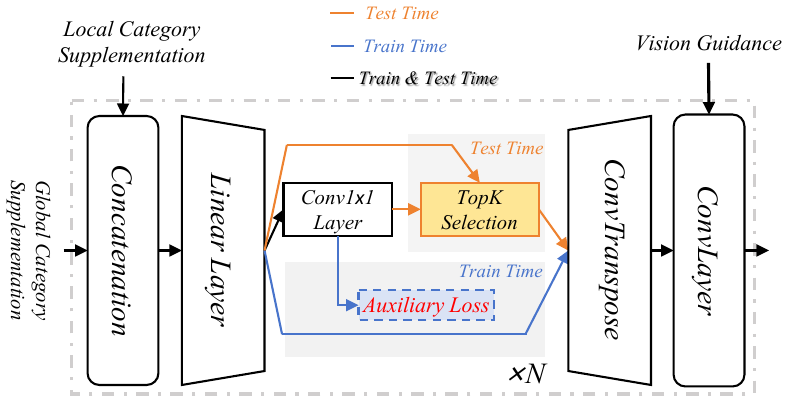}
\caption{\textbf{Decoder details.} (a) During training, the local similarity matrix is processed through concatenation, a linear layer, and a convolutional layer, with an auxiliary loss component. (b) In inference, TopK selection is applied to the local similarity matrix, followed by similar layers as in training.}
\label{fig:fig_5}
\end{figure}
\subsection{Details of The Decoder}
After the Category Supplementation Propagation, a $N_d$-layer decoder is employed to upsample the features processed by the Category Supplementation Propagation module. During this upsampling phase, an upsampled LCS $\mathcal{S}_{l}$ is integrated to supplement disturbed category boundary information within the higher-level features.

To enhance inference speed, we devised a simple yet effective acceleration technique like \cite{xie2024sed}. Specifically, as illustrated in Fig. \ref{fig:fig_5}, the fused features are first projected through a convolutional layer. During the training phase, as indicated by the blue connection line, the output $\mathbf{Y}_{\text{auxi}}$ is upsampled to compute an auxiliary loss, $\mathcal{L}_{\text{auxi}}$, defined as:
\begin{equation} 
\mathcal{L}_{\text{auxi}} = \text{CrossEntropy}(\mathbf{Y}_{\text{auxi}}, \mathbf{M}), 
\end{equation}
where $\mathbf{Y}_{\text{auxi}}$ represents the segmentation map from the auxiliary branch, and $\mathbf{M}$ denotes the ground truth labels.

In the inference phase, as shown by the orange connection line, the projected features are utilized to select the top-$k$ most significant classes. This selection reduces computational overhead, thereby optimizing inference efficiency.

\subsection{Overall Loss}
After Decoder, the final output $\mathbf{Y}$ is obtained. To perform segmentation, we employ cross-entropy as the classification loss, formulated as follows:
\begin{equation} 
\mathcal{L}_{\text{ce}} = \text{CrossEntropy}(\mathbf{Y}, \mathbf{M}), 
\end{equation}
where $\mathbf{M}$ denotes the ground truth segmentation mask.

The overall loss function in our model is defined as:
\begin{equation}
    \mathcal{L}_{\text{overall}} = \mathcal{L}_{\text{ce}} + \lambda_{\text{align}} \mathcal{L}_{\text{align}} + \lambda_{\text{auxi}} \mathcal{L}_{\text{auxi}},
\end{equation}
where \(\mathcal{L}_{\text{ce}}\) is the primary cross-entropy loss for pixel-level classification. \(\mathcal{L}_{\text{align}}\), weighted by \(\lambda_{\text{align}}\), enforces fine-grained pixel-text alignment. \(\mathcal{L}_{\text{auxi}}\), scaled by \(\lambda_{\text{auxi}}\), serves as an auxiliary term to enhance inference acceleration. The hyperparameters \(\lambda_{\text{align}}\) and \(\lambda_{\text{auxi}}\) balance these components with the main objective.

\section{Experiments And Results}
\label{experiments_results}
\begin{table*}[ht]
\caption{\textbf{Comparison with state-of-the-art methods.} The table compares performance across various methods for open-vocabulary semantic segmentation. Bold entries represent the best performance in each column, while underlined entries indicate the second-best performance.}
\label{tab:ref_0_MainTable}
\vspace{-10pt}
\footnotesize
    \begin{center}
    \resizebox{\textwidth}{!}{
    \begin{tabular}{lcccccccc}
    \toprule
        Method & VLM & Feature backbone & Training dataset & \texttt{A-847} & \texttt{PC-459} & \texttt{A-150} & \texttt{PC-59} & \texttt{PAS-20} 
        \\
        \midrule
        SPNet~\cite{xian2019semantic} & - & ResNet-101 & PASCAL VOC & - & - & - & 24.3 & 18.3  \\
        ZS3Net~\cite{bucher2019zero} & - & ResNet-101 & PASCAL VOC & - & - & - & 19.4 & 38.3 \\
        LSeg~\cite{li2022language} & ViT-B/32 & ResNet-101 & PASCAL VOC-15  & - & - & - & - & 47.4 \\
        LSeg+~\cite{ghiasi2022scaling} & ALIGN & ResNet-101 & COCO-Stuff & 2.5 & 5.2 & 13.0 & 36.0 & -  \\
        Han et al.~\cite{han2023global} & ViT-B/16 & ResNet-101& COCO Panoptic~\cite{kirillov2019panoptic} & 3.5 & 7.1 & 18.8 & 45.2 & 83.2  \\
        GroupViT~\cite{xu2022groupvit} & ViT-S/16 & - & GCC~\cite{sharma2018conceptual}+YFCC~\cite{thomee2016yfcc100m} & 4.3 & 4.9 & 10.6 & 25.9 & 50.7   \\
        ZegFormer~\cite{ding2022decoupling} & ViT-B/16 & ResNet-101 & COCO-Stuff-156 & 4.9 & 9.1 & 16.9 & 42.8 & 86.2 \\
        ZegFormer~\cite{cho2024cat} & ViT-B/16 & ResNet-101 & COCO-Stuff & 5.6 & 10.4 & 18.0 & 45.5 & 89.5  \\
        SimBaseline~\cite{xu2022simple} & ViT-B/16 & ResNet-101 & COCO-Stuff & 7.0 & - & 20.5 & 47.7 & 88.4\\
        OpenSeg~\cite{ghiasi2022scaling} & ALIGN & ResNet-101 & COCO Panoptic~\cite{kirillov2019panoptic}+LOc. Narr.~\cite{pont2020connecting} & 4.4 & 7.9 & 17.5 & 40.1 & - \\
        DeOP~\cite{han2023open} & ViT-B/16 & ResNet-101c& COCO-Stuff-156  & 7.1 & 9.4 & 22.9 & 48.8 & 91.7  \\
        PACL~\cite{mukhoti2023open} & ViT-B/16 & - & 
         GCC~\cite{sharma2018conceptual}+YFCC~\cite{thomee2016yfcc100m}  & - & - & \underline{31.4} & 50.1 & 72.3  \\
        OVSeg~\cite{liang2023open} & ViT-B/16 & ResNet-101c & COCO-Stuff+COCO Caption & 7.1 & 11.0 & 24.8 & 53.3 & 92.6  \\
        CAT-Seg~\cite{cho2024cat} & ViT-B/16 & ResNet-101 & COCO-Stuff & 8.4 & 16.6 & 27.2 & \textbf{57.5} & 93.7 \\
        SAN~\cite{xu2023side} & ViT-B/16 & -&  COCO-Stuff & 10.1 & 12.6 & 27.5 & 53.8 & \underline{94.0}   \\
        SED \cite{xie2024sed}& ConvNeXt-B & - & COCO-Stuff & \underline{11.4}  &\underline{18.6} &\textbf{31.6}  &\underline{57.3} &  94.4\\
        \rowcolor{gray!20}
        \textbf{(Ours)} & ViT-B/16 & - & COCO-Stuff & \textbf{12.0}  &\textbf{19.0} &\underline{31.4}  &\textbf{57.5} &  \textbf{95.2}\\
        \midrule
        LSeg~\cite{li2022language} & ViT-B/32 & ViT-L/16 & PASCAL VOC-15  & - & - & - & - & 52.3  \\
        OpenSeg~\cite{ghiasi2022scaling} & ALIGN & Eff-B7~\cite{tan2019efficientnet} & COCO Panoptic~\cite{kirillov2019panoptic}+LOc. Narr.~\cite{pont2020connecting} & 8.1 & 11.5 & 26.4 & 44.8 & - \\
        OVSeg~\cite{liang2023open} & ViT-L/14 & Swin-B & COCO-Stuff+COCO Caption & 9.0 & 12.4 & 29.6 & 55.7 & 94.5  \\
        Ding \textit{et al.}~\cite{ding2022open} & ViT-L/14 & - & COCO Panoptic~\cite{kirillov2019panoptic} & 8.2 & 10.0 & 23.7 & 45.9 & -  \\
        ODISE~\cite{xu2023open} & ViT-L/14 & - & COCO Panoptic~\cite{kirillov2019panoptic}& 11.1 & 14.5 & 29.9 & 57.3 & - \\
        HIPIE~\cite{wang2024hierarchical} & BERT-B~\cite{devlin2018bert} & ViT-H & COCO Panoptic~\cite{kirillov2019panoptic}& - & - & 29.0 & 59.3 & -\\
        SAN~\cite{xu2023side} & ViT-L/14 & -&  COCO-Stuff & 13.7 & 17.1 & 33.3 & 60.2 & 95.5  \\
        CAT-Seg~\cite{cho2024cat} & ViT-L/14 & Swin-B & COCO-Stuff & 10.8 & 20.4 & 31.5 & \underline{62.0} & \underline{96.6} \\
        FC-CLIP~\cite{yu2023convolutions} & ConvNeXt-L & - & COCO Panoptic~\cite{kirillov2019panoptic} & \underline{14.8} & 18.2 & 34.1 & 58.4 & 95.4  \\
        SED\cite{xie2024sed} & ViT-L & - & COCO-Stuff & 13.9 & \underline{22.6} & \underline{35.2} & 60.6 & 96.1  \\
        \rowcolor{gray!20} \textbf{(Ours)} & ViT-L/14 & - & COCO-Stuff & \textbf{16.3} & \textbf{23.9} & \textbf{37.9} & \textbf{63.4} & \textbf{97.1}  \\
        \bottomrule
    \end{tabular}
    }
    \end{center}
    \vspace{-10pt}
\end{table*}

\subsection{Datasets}
This study utilizes the comprehensive COCO-Stuff dataset to train the model. COCO-Stuff includes approximately 118,000 images with dense annotations across 171 unique semantic categories. With a model trained on this dataset, we perform evaluations on several widely used semantic segmentation benchmarks to validate the effectiveness of our proposed method and benchmark it against state-of-the-art approaches from existing literature.
The ADE20K dataset serves as a large-scale benchmark for semantic segmentation, comprising 20,000 training images and 2,000 validation images. In the context of open-vocabulary semantic segmentation, ADE20K provides two distinct test sets: A-150, which includes 150 frequently occurring categories, and A-847, which spans 847 categories.
The PASCAL VOC dataset, one of the early resources for object detection and segmentation, offers around 1,500 training images and an equal number of validation images, covering 20 different object classes. For open-vocabulary semantic segmentation, this dataset is referred to as PAS-20.
An extension of the original PASCAL VOC dataset, the PASCAL-Context dataset is specifically designed for semantic segmentation tasks. In the open-vocabulary segmentation setting, it features two separate test sets: PC-59, which includes 59 categories, and PC-459, covering 459 categories.

\subsection{Experimental Details}
All our experiments were conducted on 8 NVIDIA 4090 GPUs, with a batch size of 1 per GPU. The learning rate was set to 0.0002, with a maximum of 80,000 training iterations, and testing was performed every 5,000 iterations. The hyperparameters were set as $\lambda_{auxi} = 0.2$, $\lambda_{align} = 0.02$, with the default values of $N = 1$ and $N_d = 3$. For the backbone, we utilized \texttt{ViT-B/16} and \texttt{ViT-L/14@336px}. The training approach primarily involved fine-tuning the attention layers. Regarding the text prompt template, we employed the simple format "a photo of $<CLASS>$". For vision guidance, we utilize the CLIP features from layers 7 and 15 for ViT-L/14@336px, and from layers 3 and 7 for ViT-B/16.

\subsection{Comparisons with State-of-the-Art Methods}

In Table~\ref{tab:ref_0_MainTable}, we conduct a detailed comparison between our proposed method and several state-of-the-art approaches across five widely-used open-vocabulary semantic segmentation benchmarks: A-847, PC-459, A-150, PC-59, and PAS-20. The data is adapted from \cite{xie2024sed}. These datasets encompass diverse scenes and object categories, providing a comprehensive evaluation for open-vocabulary models. Our method consistently outperforms the competing approaches across all datasets, underscoring its robustness and adaptability to different segmentation challenges.

Using the ViT-B/16 backbone, our method achieves an mIoU of 12.0 on A-847, 19.0 on PC-459, 31.4 on A-150, 57.5 on PC-59, and 95.2 on PAS-20. In comparison, other models with similar backbones and training configurations, such as CAT-Seg~\cite{cho2024cat} and OVSeg~\cite{liang2023open}, achieve lower mIoU scores on these datasets. Specifically, CAT-Seg achieves only 8.4 mIoU on A-847 and 16.6 on PC-459, while OVSeg records 24.8 on A-150 and 53.3 on PC-59. These results demonstrate our model's superior feature representation capabilities, particularly in open-vocabulary contexts where the model must generalize to unseen categories. Our method's effective generalization is likely attributed to its improved pixel-level alignment module, which minimizes semantic distortion during segmentation, and its advanced fusion mechanisms that enhance feature discrimination without overfitting to specific categories.

When scaling up to the larger ViT-L/14 backbone, our method further solidifies its advantage, achieving an mIoU of 16.3 on A-847, 23.9 on PC-459, 37.7 on A-150, 63.1 on PC-59, and 96.9 on PAS-20. This represents a notable improvement compared to ViT-B/16 and highlights the scalability of our approach. Importantly, these performance gains are achieved without relying on external datasets or auxiliary backbones, in contrast to other top-performing models like CAT-Seg~\cite{cho2024cat} and OVSeg, which often require additional labeled data or complex model architectures. OVSeg, for example, incorporates additional data and a multi-stage training process, yet achieves comparable or even lower scores on PC-459 and PAS-20 relative to our method, suggesting that our approach provides a more data-efficient and computationally feasible solution.
The advantages of our method are particularly prominent on the PC-59 and PAS-20 test sets, where segmentation tasks are more challenging due to high variability in object appearance, scale, and context. Our model achieves 63.1 mIoU on PC-59 and 96.9 on PAS-20, substantially outperforming models such as SAN~\cite{xu2023side}, which utilizes extensive external datasets to enhance segmentation precision but records a lower mIoU on PC-59. This indicates that our model can handle complex, high-variation datasets more effectively, likely due to its balanced design that enhances feature richness without introducing excessive complexity. The integration of our efficient multimodal fusion layer contributes to improved segmentation accuracy in scenes with intricate visual contexts by adaptively combining visual cues and textual embeddings, thus enabling the model to identify object boundaries with higher precision.

\subsection{Ablation Experiments}
\subsubsection{Ablation Setting}
\quad

In our ablation experiments, the default parameters chosen are as follows: the kernel size is set to 3, \(\gamma\) is trainable, the value of \(\lambda_{\text{align}}\) is set to 0.02, and the number of layers \(N_d\) in the P2Transformer is set to 1. We conduct ablation studies and parameter optimization based on these settings.

\subsubsection{Impact of Integrating Different Components}
\begin{table*}[t]
\caption{\textbf{Impact of different modules in FGAseg.} The table shows the performance improvements with the integration of different modules into the baseline.}
    \label{tab:ref_1_MainComp}
    \centering
    \resizebox{0.7\linewidth}{!}{
   \begin{tabular}{ccc|ccccc}
        \toprule
         P2Tformer & T2Ploss & Local Cost & \texttt{A-847} & \texttt{PC-459} & \texttt{A-150} & \texttt{PC-59} & \texttt{PAS-20} 
         \\
        \midrule
        & &  & 11.1 & 18.1 & 29.9 & 55.9 & 93.8 \\
        \checkmark & &  & 11.6 & 18.4 & 30.3 & 56.6 & 94.0 \\
        \checkmark & \checkmark & & 12.1 & 18.3 & 31.4 & 56.3 & 94.9 \\
        \rowcolor{gray!15} \checkmark & \checkmark & \checkmark & 12.0 & 19.0 & 31.4 & 57.5 & 95.2\\
        \bottomrule
\end{tabular}}%
\end{table*}
\quad

Table \ref{tab:ref_1_MainComp} demonstrates the effects of integrating different components into the FGAseg model. The baseline performance with only the P2Tformer indicates a solid starting point. Incorporating the T2Ploss results in slight improvements, suggesting enhanced segmentation capabilities.
The addition of the Local Cost component yields the best performance, highlighting its significant contribution to refining feature representation and accuracy. Overall, these findings underscore the importance of each integrated component in enhancing the model's performance in semantic segmentation tasks.

\begin{figure*}[ht]
\centering
\includegraphics[width=0.85\linewidth]{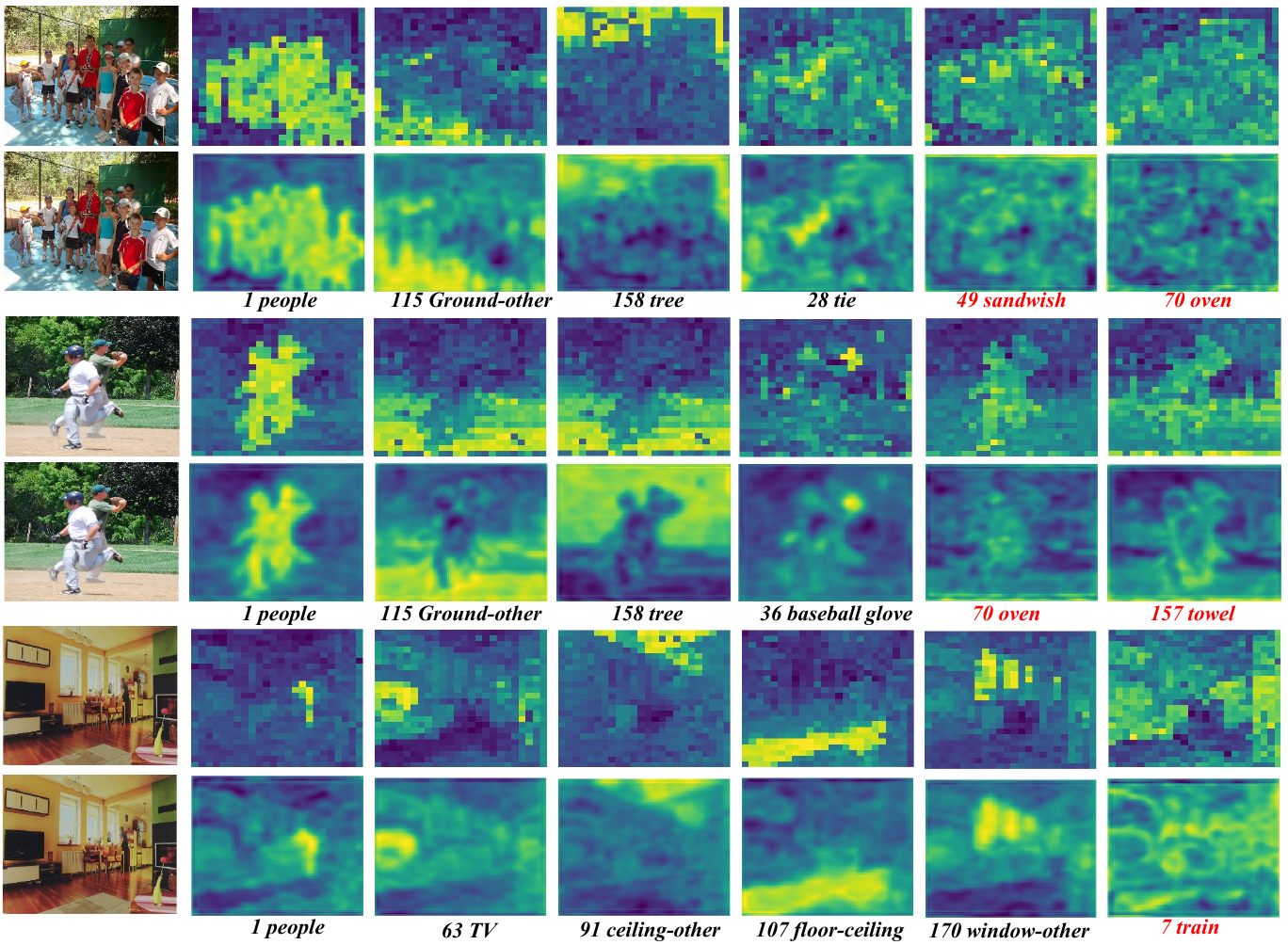}
\caption{\textbf{Visualization of pseudo-masks.} The first row represents global category supplementation (24×24 resolution), while the second row represents local category supplementation (96×96 resolution). We illustrate several successful pseudo label masks (in \textbf{black}) and cases that theoretically should not match correctly (in \textbf{\textcolor{red}{red}}). All the pictures are from COCO-val dataset. The category indices follow the COCO dataset.}
\label{fig:fig_6}
\end{figure*}
\subsubsection{Exploration of The Pixel-Text Alignment Module}
\begin{table}[t]
\centering
% \caption{\textbf{Ablation study on different designs in Pixel Alignment Module.}}
\caption{\textbf{Ablation study on different designs in Pixel Alignment Module.} The table presents the performance of various configurations in the Pixel Alignment Module, including the effects of different components like the P2Tformer, $N_d$, and $\gamma$.}
\label{tab:ref_2_Pixel2TextAlignModule}
\resizebox{\linewidth}{!}{
\begin{tabular}{c|c|cccccc}
    \toprule
        \multicolumn{7}{l}{\textbf{P2TFormer}} \\ 
    \midrule 
    \multirow{3}{*}{(a)} & Frozen CLIP & \texttt{A-847} & \texttt{PC-459} & \texttt{A-150} & \texttt{PC-59} & \texttt{PAS-20} \\
    \cmidrule{2-7}
        & w/o P2Tformer & 8.5 & 14.8 & 25.6 & 52.1 & 93.6 \\
        & \cellcolor{gray!15} w/ P2Tformer & \cellcolor{gray!15}9.0 & \cellcolor{gray!15}15.1 & \cellcolor{gray!15}25.2 & \cellcolor{gray!15}53.3 & \cellcolor{gray!15}94.0 \\
    \midrule
    
  \multirow{9}{*}{(b)} & $N_d$ & \texttt{A-847} & \texttt{PC-459} & \texttt{A-150} & \texttt{PC-59} & \texttt{PAS-20} \\
    \cmidrule{2-7}
    & \cellcolor{gray!15}1 & \cellcolor{gray!15}12.0 & \cellcolor{gray!15}19.0 & \cellcolor{gray!15}31.4 & \cellcolor{gray!15}57.5 & \cellcolor{gray!15}95.2 \\ 
    & 2 & 11.6 & 19.2 & 31.4 & 57.6 & 94.8 \\ 
    & 3 & 12.1 & 18.9 & 31.5 & 56.9 & 95.1 \\ 
    & 4 & 11.7 & 19.0 & 31.4 & 57.6 & 95.2 \\ 
    & 5 & 11.7 & 19.2 & 31.3 & 57.0 & 95.0 \\ 
    & 6 & 12.0 & 18.8 & 31.5 & 57.4 & 94.9 \\ 
    & 7 & 11.9 & 19.0 & 31.1 & 56.9 & 94.8 \\
    \midrule
    
  \multirow{8}{*}{(c)} & $\gamma$ & \texttt{A-847} & \texttt{PC-459} & \texttt{A-150} & \texttt{PC-59} & \texttt{PAS-20} \\
    \cmidrule{2-7}
    & \cellcolor{gray!15}Trainable & \cellcolor{gray!15}12.0 & \cellcolor{gray!15}19.0 & \cellcolor{gray!15}31.4 & \cellcolor{gray!15}57.5 & \cellcolor{gray!15}95.2 \\
    & 0.01 & 11.9 & 18.9 & 31.6 & 56.9 & 95.1 \\
    & 0.1 & 12.1 & 19.2 & 31.6 & 57.0 & 95.1 \\
    & \cellcolor{gray!15}0.5 & \cellcolor{gray!15}12.1 & \cellcolor{gray!15}19.1 & \cellcolor{gray!15}31.6 & \cellcolor{gray!15}57.8 & \cellcolor{gray!15}95.1 \\
    & 1 & 11.7 & 19.1 & 31.5 & 57.3 & 95.4 \\
    & 5 & 11.7 & 19.0 & 31.1 & 57.4 & 95.0 \\
    \midrule 
    \multicolumn{7}{l}{\textbf{T2PLoss}} \\ 
    \midrule 
  \multirow{10}{*}{(d)} & $\lambda_{align}$ & \texttt{A-847} & \texttt{PC-459} & \texttt{A-150} & \texttt{PC-59} & \texttt{PAS-20} \\
    \cmidrule{2-7}
    & 0.002 & 12.0 & 18.9 & 31.6 & 57.3 & 95.0 \\
    & 0.005 & 11.6 & 19.1 & 31.2 & 57.3 & 95.1 \\
    & \cellcolor{gray!15}0.02 & \cellcolor{gray!15}12.0 & \cellcolor{gray!15}19.0 & \cellcolor{gray!15}31.4 & \cellcolor{gray!15}57.5 & \cellcolor{gray!15}95.2 \\ 
    & 0.2 & 11.7 & 19.2 & 31.3 & 57.6 & 95.2 \\
    & 0.5 & 11.8 & 19.2 & 30.8 & 57.3 & 95.3 \\
    & 1 & 11.3 & 18.2 & 30.3 & 57.3 & 95.2 \\
    & 5 & 9.3 & 16.9 & 27.9 & 57.5 & 94.9 \\

    \bottomrule
\end{tabular}}
\vspace{-5pt}

\vspace{-10pt}
\end{table}

\quad

\textbf{The Effectiveness of Pixel-Text Alignment.}
The addition of P2Tformer substantially enhances the alignment between vision and text features in pixel level compared to the baseline frozen CLIP model, which lacks this component. The module achieves higher scores across all datasets, affirming that pixel-text alignment contributes to finer-grained feature alignment and improved segmentation outcomes. Specifically, by introducing an alignment mechanism dedicated to pixel-text correspondence, P2Tformer provides a specialized layer of abstraction that reduces discrepancies in the vision and text embeddings, ensuring a more synchronized feature space. This alignment has proven effective in tackling the challenges of complex segmentation tasks, as shown by its performance improvements over the baseline, thereby demonstrating the critical role of precise pixel-text alignment in OVS.

\textbf{Comparison of Different Layers of P2Tformer.}
Varying the depth \(N_d\) of P2Tformer reveals that layer depth significantly impacts alignment effectiveness and computational efficiency. Results indicate that both \(N_d = 1\) and \(N_d = 6\) yield high performance, with \(N_d = 1\) offering an optimal trade-off between computational cost and alignment efficacy. This finding suggests that a shallow P2Tformer layer can capture the essential pixel-text interaction patterns, while deeper configurations marginally increase alignment at the expense of efficiency. Notably, as the layer depth increases beyond \(N_d = 1\), the performance gains become less pronounced, indicating diminishing returns. This suggests that for applications requiring fast inference times, a single-layer P2Tformer may be preferable, balancing both effectiveness and speed.

\textbf{The Effectiveness of Trainable \(\gamma\) in P2Tformer.}
Introducing a trainable \(\gamma\) parameter consistently outperforms fixed values across all datasets, suggesting that dynamic adjustment of this scaling factor enhances alignment quality. Allowing \(\gamma\) to adapt to the data in each training scenario improves the model's adaptability and robustness. Specifically, while a fixed \(\gamma\) such as \(\gamma = 0.5\) yields stable performance, a learnable \(\gamma\) achieves finer control over feature scaling, facilitating enhanced feature alignment. This adaptability is particularly beneficial for models handling diverse and complex visual-textual pairs, as the ability to adjust \(\gamma\) based on feature relevance directly contributes to superior alignment outcomes.

\textbf{The Impact of \(\lambda_{align}\) for \(\mathcal{L}_{align}\).}
Modulating \(\lambda_{align}\) in the P2Tloss (\(\mathcal{L}_{align}\)) has a substantial effect on performance, where an intermediate value of \(\lambda_{align} = 0.02\) optimally balances the alignment across datasets. This balanced value enables strong feature correspondence without over-penalizing misalignment, which could otherwise disrupt the primary learning objectives of the model. Variations in \(\lambda_{align}\) show that while smaller values do not impose sufficient alignment constraints, larger values can overly dominate the loss function, leading to reduced flexibility in feature learning. Hence, a carefully tuned \(\lambda_{align}\) ensures that alignment is adequately prioritized without overwhelming other aspects of model learning.

\subsubsection{Exploration of The Local Alignment Cost}
\begin{table}[ht]
\caption{\textbf{Exploration of Different Design Local Alignment Cost.} This table investigates the effect of kernel size and kernel normalization on local alignment cost across several datasets: \texttt{A-847}, \texttt{PC-459}, \texttt{A-150}, \texttt{PC-59}, and \texttt{PAS-20}.}
    \label{tab:ref_3_LocalCost}
    \centering
    \resizebox{\linewidth}{!}{
    \begin{tabular}{l|c|ccccc}
        \toprule
        \multirow{8}{*}{(a)} &  Kernel Size &  \texttt{A-847} &  \texttt{PC-459} &  \texttt{A-150} &  \texttt{PC-59} &  \texttt{PAS-20} \\
        \cmidrule{2-7}
        &  1 &  11.8 &  18.9 &  31.4 &  57.3 &  95.1 \\
        & 3 &  12.0 &  19.0 &  31.4 &  57.5 &  95.2 \\
        &  5 &  11.9 &  19.0 &  31.3 &  57.3 &  95.4 \\
        &  7 &  11.9 &  19.3 &  31.8 &  57.3 &  95.2 \\
        & \cellcolor{gray!15}9 & \cellcolor{gray!15}12.0 & \cellcolor{gray!15}19.4 & \cellcolor{gray!15}31.9 & \cellcolor{gray!15}57.4 & \cellcolor{gray!15}95.4 \\
        & \cellcolor{gray!15}11 & \cellcolor{gray!15}12.1 & \cellcolor{gray!15}19.4 & \cellcolor{gray!15}32.0 & \cellcolor{gray!15}57.2 & \cellcolor{gray!15}95.2\\
        & 13 & 12.0 & 18.9 & 31.8 & 57.4 & 95.0\\
        & 15 & 11.9 & 18.9 & 31.2 & 57.2 & 95.1  \\
        \midrule
        \multirow{4}{*}{(b)} &  Kernel Norm &  \texttt{A-847} &  \texttt{PC-459} &  \texttt{A-150} &  \texttt{PC-59} &  \texttt{PAS-20} \\
        \cmidrule{2-7}
        &  No &  10.7 &  18.5 &  29.4 &  56.7 &  94.8 \\
        &  \cellcolor{gray!15}Yes &  \cellcolor{gray!15}12.0 &  \cellcolor{gray!15}19.0 &  \cellcolor{gray!15}31.4 &  \cellcolor{gray!15}57.5 &  \cellcolor{gray!15}95.2 \\
        \bottomrule
    \end{tabular}
    }
\end{table}

\quad

\textbf{The Impact of Kernel Size in Local Alignment Cost.}
Table \ref{tab:ref_3_LocalCost}(a) presents the effect of various kernel sizes on model performance across different datasets. A kernel size of 9 consistently achieves the highest mean Intersection over Union (mIoU), while sizes 11 and 5 follow closely in performance. These results suggest that larger kernels enhance the receptive field, allowing the model to capture more contextual information, which is beneficial for detailed segmentation. Beyond a kernel size of 9, performance stabilizes, indicating diminishing returns with further size increases. This stabilization reflects that once a sufficient amount of context is incorporated, additional kernel size may not further improve feature extraction or segmentation accuracy. Consequently, a kernel size of 9 provides an optimal balance, maximizing performance gains without unnecessary computational overhead.

\textbf{The Impact of Kernel Normalization in Local Alignment Cost.}
Table \ref{tab:ref_3_LocalCost}(b) assesses the role of kernel normalization on model performance. The results demonstrate that applying kernel normalization leads to a significant improvement in mIoU across all datasets, as compared to no normalization. This improvement highlights the importance of normalization in refining feature representation and stabilizing model predictions. By normalizing kernel outputs, the model achieves a more balanced alignment cost, which improves its ability to handle varying scales and contrasts within images. This finding underscores kernel normalization as an essential step in optimizing the local alignment cost, contributing to more consistent and accurate segmentation.

\begin{figure*}[ht]
\centering
\includegraphics[width=0.85\linewidth]{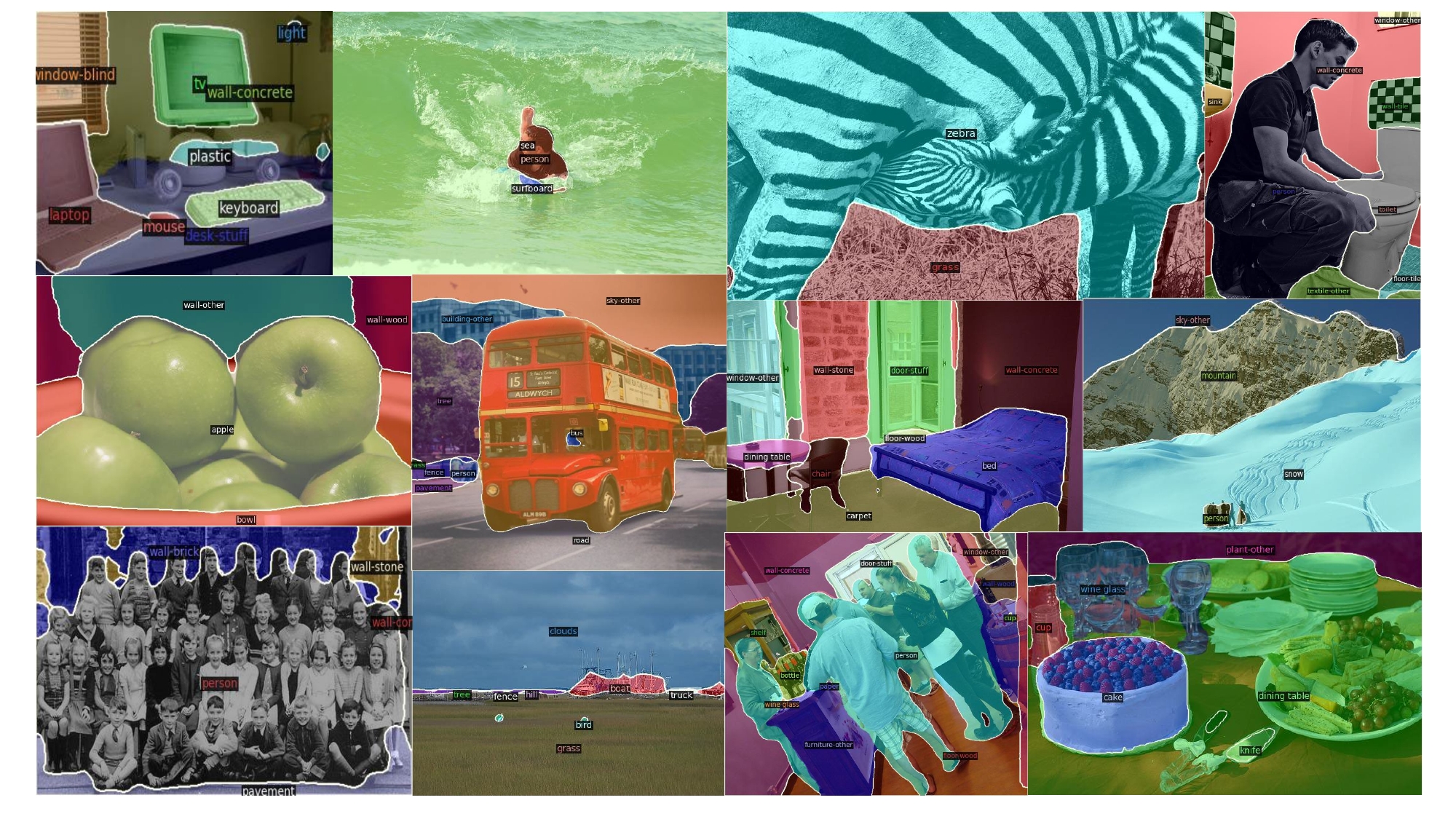}
\includegraphics[width=0.85\linewidth]{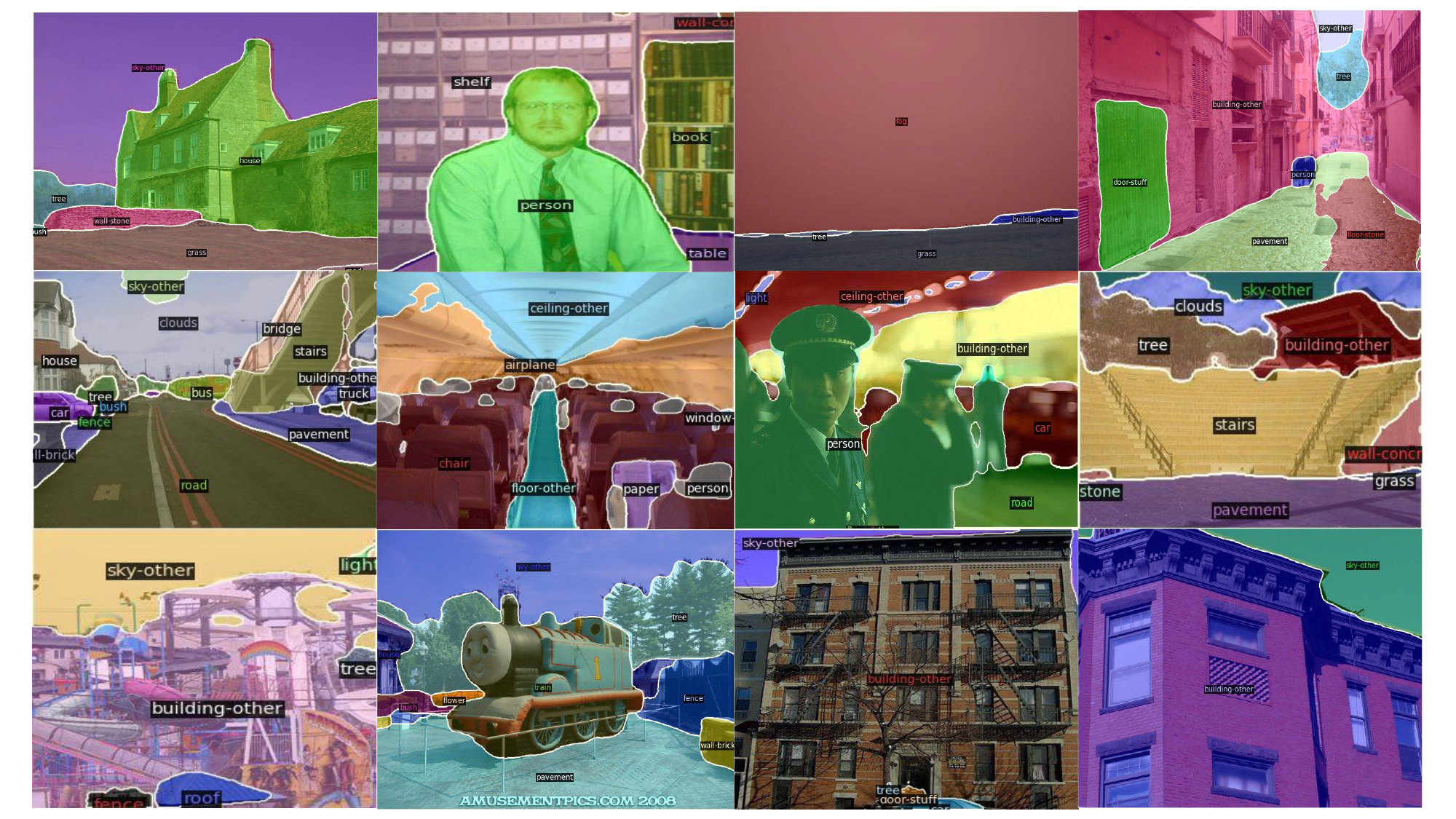}
\caption{\textbf{Segmentation case visualization.} The data is sourced from the COCO dataset and the ADE20k dataset. None of the datasets were seen by the model during training, and the data labels are consistent with the respective datasets.}
\label{fig:fig_78}
\end{figure*}

\subsubsection{Effectiveness of The Inference Acceleration}
\quad

Table \ref{tab:ref_4_Decoder} compares various decoder configurations. The "NoFast+Cat" design achieves the highest mIoU, highlighting its effectiveness in preserving feature integrity. In contrast, the "Fast+Add" configuration experiences a slight drop in performance while maintaining comparable time efficiency. This analysis underscores the importance of selecting appropriate decoder strategies to effectively balance accuracy and efficiency in semantic segmentation tasks.

In Table \ref{tab:ref_5_TopN}, we observe that as the Top-N value increases, the mIoU remains relatively stable, indicating consistent performance across different configurations. The time metrics show an expected increase, confirming the trade-off between accuracy and inference speed. Notably, a Top-N of 1 delivers competitive mIoU with minimal computation time, suggesting its suitability for real-time applications.

\begin{table}[ht]
\caption{\textbf{Ablation study on different designs in Decoder.} The table compares the performance of various decoding strategies.}
\label{tab:ref_4_Decoder}
\centering
\resizebox{\linewidth}{!}{
\begin{tabular}{c|c|cccccc}
    \toprule
 \multirow{6}{*}{(a)} &  Decoder & \texttt{A-847} & \texttt{PC-459} & \texttt{A-150} & \texttt{PC-59} & \texttt{PAS-20} 
     \\
    \cmidrule{2-7}
      &  Fast+Cat & 11.7 & 18.8 & 31.6 & 57.5 & 94.6 \\ 
      &  Fast+Add & 12.0 & 18.8 & 31.6 & 57.1 & 94.7 \\ 
     \rowcolor{gray!15} &  NoFast+Cat & 12.0 & 19.0 & 31.4 & 57.5 & 95.2 \\ 
      &  NoFast+Add & 11.8 & 19.1 & 30.9 & 57.4 & 95.2 \\ 
    \bottomrule
\end{tabular}}
\end{table}

\begin{table}[ht]
\centering
\caption{\textbf{Ablation study on different designs in Top-N for mIoU and Time.} This table evaluates the impact of varying Top-N values on the mIoU and processing time across five datasets.}
\label{tab:ref_5_TopN}
\resizebox{\linewidth}{!}{
\begin{tabular}{c|c|ccccc}
    \toprule
    Top-N & Metric & \texttt{A-847} & \texttt{PC-459} & \texttt{A-150} & \texttt{PC-59} & \texttt{PAS-20} \\
    \cmidrule{1-7}
    \multirow{2}{*}{1} & mIoU \cellcolor{gray!15}& 11.9 \cellcolor{gray!15}& 19.0 \cellcolor{gray!15}& 31.4 \cellcolor{gray!15}& 57.5 \cellcolor{gray!15}& 95.2 \cellcolor{gray!15}\\ 
      & Time \cellcolor{gray!15}& 219  \cellcolor{gray!15}& 189  \cellcolor{gray!15}& 146  \cellcolor{gray!15}& 126  \cellcolor{gray!15}& 118 \cellcolor{gray!15} \\
    \midrule       
    \multirow{2}{*}{4} & mIoU & 12.0 & 19.0 & 31.5 & 57.5 & 95.2 \\ 
      & Time & 327  & 274  & 185  & 150  & 127  \\
    \midrule
    \multirow{2}{*}{8} & mIoU & 12.0 & 19.0 & 31.4 & 57.5 & 95.2 \\
      & Time & 399  & 345  & 210  & 164  & 131  \\
    \midrule
    \multirow{2}{*}{16} & mIoU & 12   & 19   & 31.5 & 57.5 & 95.2 \\ 
      & Time & 256  & 419  & 240  & 179  & 132  \\
    \midrule
    \multirow{2}{*}{32} & mIoU & 12.0 & 18.9 & 31.5 & 57.5 & 95.2 \\ 
      & Time & 487  & 465  & 293  & 186  & 132  \\
    \bottomrule
\end{tabular}}

\end{table}

\begin{table}[t]
\centering
\label{tab:ref_6_TrainStrategy}
\caption{\textbf{Ablation study on different designs in Decoder.} The table compares the performance of different decoding strategies (\texttt{Prompt-16}, \texttt{Attention}, and \texttt{Full-tuning}) across five datasets.}
\resizebox{\linewidth}{!}{
\begin{tabular}{c|ccccc}
    \toprule
Strategy & \texttt{A-847} & \texttt{PC-459} & \texttt{A-150} & \texttt{PC-59} & \texttt{PAS-20} 
    \\
    \cmidrule{1-6}
   Prompt-16 & 9.2 & 15.4 & 26.2 & 54.2 & 93.2 \\
  \cellcolor{gray!15}  Attention & \cellcolor{gray!15}12.0 & \cellcolor{gray!15}19.0 & \cellcolor{gray!15}31.4 & \cellcolor{gray!15}57.5 & \cellcolor{gray!15}95.2 \\
   Full-tuning & 11.0 & 18.5 & 29.0 & 57.8 & 95.4 \\
    \bottomrule
\end{tabular}}
\vspace{-5pt}
\end{table}

\subsubsection{Effectiveness of The Different Training Strategy}
\quad

Table VII illustrates the impact of different training strategies on model performance across various datasets. The results demonstrate that the \texttt{Attention} strategy consistently achieves the highest mean Intersection over Union (mIoU) across all datasets, indicating its effectiveness in enhancing feature learning and alignment. In comparison, the \texttt{Prompt-16} strategy yields the lowest mIoU, highlighting its limited capability in capturing complex feature interactions.
Although the \texttt{Full-tuning} strategy provides competitive results, it falls short of matching the performance achieved by the \texttt{Attention} method. This contrast suggests that the \texttt{Attention} approach is more effective at optimizing the model’s capacity for nuanced feature extraction without extensive parameter updates, as is necessary in \texttt{Full-tuning}. These findings emphasize the crucial role of the chosen training strategy, especially the \texttt{Attention} approach, in boosting segmentation performance by improving spatial and contextual awareness.

\subsection{Segmentation Visulizations}

\subsubsection{Visualization of pseudo-masks} 
In Fig. \ref{fig:fig_6}, we present a visualization of the pseudo-masks. These pseudo-masks serve as supplementary category boundary during forward propagation. We illustrate the pseudo-masks' ability to enhance class boundaries, demonstrating, for instance, how the model provides localization and essential boundary information for categories such as "people," which is crucial during forward propagation. However, for certain categories absent in the images, the model does not generate corresponding boundary information, indicating its ability to discern between categories to some extent.

\subsubsection{Segmentation Result Comparison.}
Table \ref{fig:fig_78} illustrate distinct object boundaries and successful labeling of multiple classes, ranging from common indoor items like "laptop" and "keyboard" to outdoor entities such as "zebra" and "bus." The model demonstrates robust performance across diverse scenes, including crowded environments and natural landscapes, maintaining clarity in edge definition and semantic accuracy. Notable strengths include the precise delineation of fine-grained objects (e.g., "mouse" and "surfboard") and the consistent recognition of large-scale regions like "sky" and "mountain." However, minor mislabeling or blending of adjacent regions suggests scope for enhancement in handling overlapping objects or complex textures. Overall, the results reflect a balanced trade-off between accuracy and generalization across varying conditions.
\section{Conclusion}
\label{sec:conclusion}
In this paper, we presented the \texttt{FGASeg} framework, a novel approach designed to address the challenges of open-vocabulary segmentation by achieving fine-grained pixel-level alignment between vision and text. Our proposed Pixel-Text Alignment Transformer (P2Tformer) and Text-Pixel Alignment Loss (T2Ploss) enable precise pixel-text alignment while preserving the pretrained image-text alignment of vision-language models (VLMs). Additionally, the Category Supplementation Propagation module leverages cosine and convolution-based similarity matrices as pseudo-masks, enriching category boundary information and enhancing segmentation performance.
Extensive experiments demonstrate that \texttt{FGASeg} achieves competitive results across multiple benchmark datasets, highlighting its ability to combine pixel-level alignment with robust boundary information for effective semantic segmentation. Future work will explore further improvements in cross-modal alignment and extend the framework to additional multimodal tasks.

% \newpage
\begin{IEEEbiography}[{\includegraphics[width=1in,height=1.25in,clip,keepaspectratio]{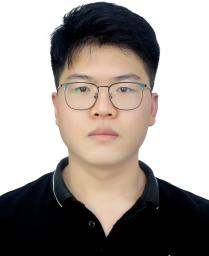}}]{Bingyu Li}
 is pursuing the Ph.D. degree in the School of Information Science and Technology, University of Science and Technology of China, Hefei, Anhui, China. His research interests include Zero-shot learning and computer vision.
\end{IEEEbiography}

\vspace{-33pt}
\begin{IEEEbiography}[{\includegraphics[width=1in,height=1.25in,clip,keepaspectratio]{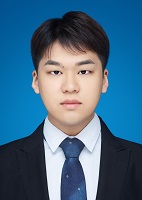}}]{Da Zhang}
is directly pursuing the Ph.D. degree in the School of Artificial Intelligence, Optics and Electronics (iOPEN), Northwestern Polytechnical University, Xi’an, China. His research interests include co mputer vision andpattern recognition.
\end{IEEEbiography}

\vspace{-22pt}
\begin{IEEEbiography}[{\includegraphics[width=1in,height=1.25in,clip,keepaspectratio]{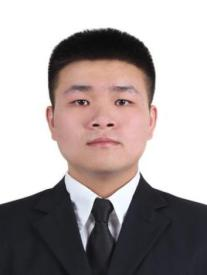}}]{Zhiyuan Zhao}
received the B.E. degree and the Ph.D. degree in computer science and technology from the Northwestern Polytechnical University, Xi’an 710072, Shaanxi, P. R. China, in 2018 and 2023 respectively. He is currently with the Institute of Artificial Intelligence (TeleAI), China Telecom, P. R. China.
\end{IEEEbiography}

\vspace{-22pt}
\begin{IEEEbiography}[{\includegraphics[width=1in,height=1.25in,clip,keepaspectratio]{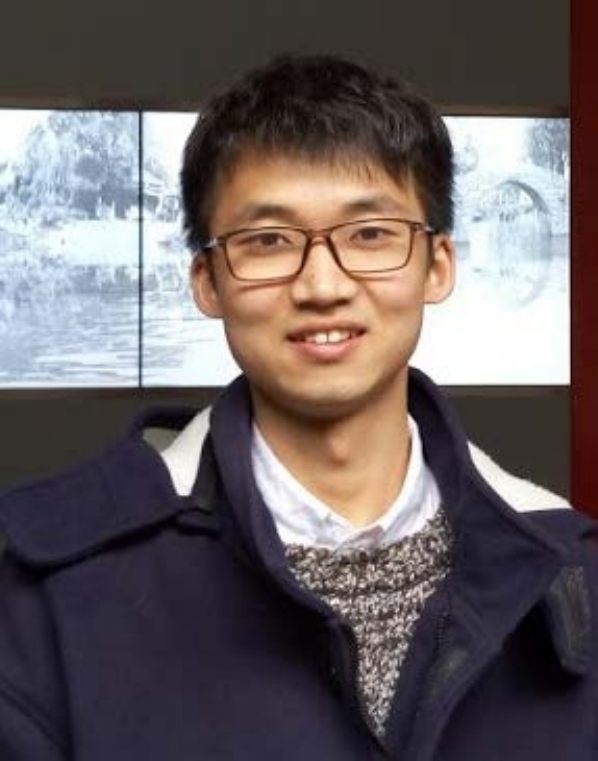}}]{Junyu Gao}
received the B.E. degree and the Ph.D. degree in computer science and technology from the Northwestern Polytechnical University, Xi’an 710072, Shaanxi, P. R. China, in 2015 and 2021 respectively. He is currently an associate professor with the School of Artificial Intelligence, Optics and
Electronics (iOPEN), Northwestern Polytechnical University, Xi’an, China. His research interests include computer vision and pattern recognition.
\end{IEEEbiography}

\vspace{-22pt}
\begin{IEEEbiography}[{\includegraphics[width=1in,height=1.25in,clip,keepaspectratio]{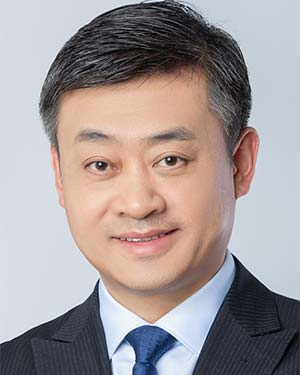}}]{Xuelong Li}
is the CTO and Chief Scientist of China Telecom, where he founded the Institute of Artificial Intelligence (TeleAI) of China Telecom. 
\end{IEEEbiography}

\vfill

\end{document}